\documentclass[10pt,twocolumn,letterpaper]{article}

\usepackage[pagenumbers]{cvpr} 

\usepackage{graphicx}
\usepackage{amsmath}
\usepackage{amssymb}
\usepackage{booktabs}
\usepackage{comment}
\usepackage{graphicx}
\usepackage{wrapfig}
\usepackage{pgfplots}
\usepackage{tikzscale}
\pgfplotsset{compat=1.18}
\usepackage{longtable,booktabs} 
\usepackage{subcaption}
\usepackage{float}
\usepackage[accsupp]{axessibility}
\usepackage[pagebackref,breaklinks,colorlinks]{hyperref}

\usepackage[capitalize]{cleveref}
\crefname{section}{Sec.}{Secs.}
\Crefname{section}{Section}{Sections}
\Crefname{table}{Table}{Tables}
\crefname{table}{Tab.}{Tabs.}
\newcommand*\samethanks[1][\value{footnote}]{\footnotemark[#1]}

\def\cvprPaperID{9} 
\def\confName{NFVLR@CVPR }
\def\confYear{2023}

\begin{document}

\title{Abstract Visual Reasoning Enabled by Language}

\author{
Giacomo Camposampiero\thanks{Equal Contribution}\and Loïc Houmard\samethanks 
 \and Benjamin Estermann \and Joël Mathys \and Roger Wattenhofer\\
 ETH Zürich, Switzerland\\
 {\tt\small \{gcamposampie, lhoumard, estermann, jmathys, wattenhofer\}@ethz.ch}
}
\maketitle

\begin{abstract}
    While artificial intelligence (AI) models have achieved human or even superhuman performance in many well-defined applications, they still struggle to show signs of broad and  flexible intelligence.
    The Abstraction and Reasoning Corpus (ARC), a visual intelligence benchmark introduced by François Chollet, aims to assess how close AI systems are to human-like cognitive abilities.
    Most current approaches rely on carefully handcrafted domain-specific program searches to brute-force solutions for the tasks present in ARC.
    In this work, we propose a general learning-based framework for solving ARC. It is centered on transforming tasks from the vision to the language domain.
    This composition of language and vision allows for pre-trained models to be leveraged at each stage, enabling a shift from handcrafted priors towards the learned priors of the models. 
    While not yet beating state-of-the-art models on ARC, we demonstrate the potential of our approach, for instance, by solving some ARC tasks that have not been solved previously.
    
\end{abstract}

\section{Introduction}
\begin{figure}[b]
        \begin{center}
        \includegraphics[width=6cm]{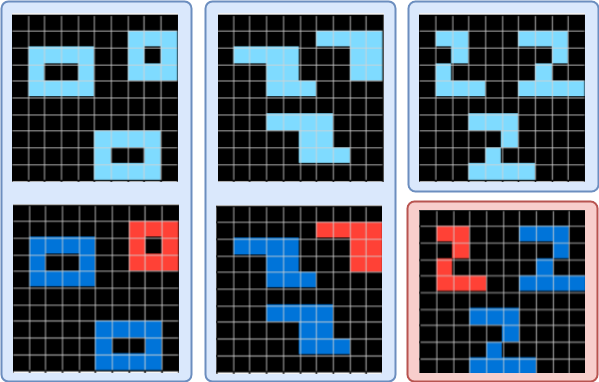}
        \vspace{-10pt}
        \end{center}
        \caption{Each ARC task consists of pairs of input and output images that describe the task. To solve the task, the missing output image corresponding to the given test input must be predicted.}
        \label{fig:arc_task}
    \end{figure}

    The Abstraction and Reasoning Corpus (ARC), introduced by Chollet~\cite{chollet12019}, is designed to measure the progress of artificial intelligence (AI) by testing the ability to learn and adapt to unseen tasks.
    ARC consists of a small dataset of only 1000 tasks similar to the example in Figure \ref{fig:arc_task}. For each task, two to six demonstrations are given and the solution of a given test input has to be predicted.  
    By design, the number of tasks is very limited, which makes it especially challenging for learning based approaches to be successful on ARC. Usually, state of the art models have to rely on enormous amounts of data to correctly estimate and learn the underlying functions directly from the given data.

    Learning only from the provided tasks to solve ARC is most likely not sufficient. As argued by Chollet~\cite{chollet12019}, prior knowledge closely aligned to human intelligence should be incorporated into the approaches.
    The most important priors cover objectness, goal-directness, numbers and counting as well as elementary geometry \cite{chollet12019, Spelke2007-rm}.
    These include the concept of objects, that they can persist, comparing sizes or checking for symmetries.
    The main challenge lies in how to incorporate these priors in an approach to solve ARC.

    One can make them explicit through handcrafted heuristics, for example with domain-specific languages (DSLs)~\cite{IceCuber,Kaggle2,Kaggle3}.
    These DSLs can then be used to brute-force programs that search and propose correct solutions.
    However, it is extremely difficult to capture all important priors explicitly and apply them correctly.
    Further, an exhaustive search is computationally expensive and might even yield wrong results, as not all combinations of priors are equally likely to appear at the same time.

    \begin{figure*}[t]
        \begin{center}
        \includegraphics[width=17cm]{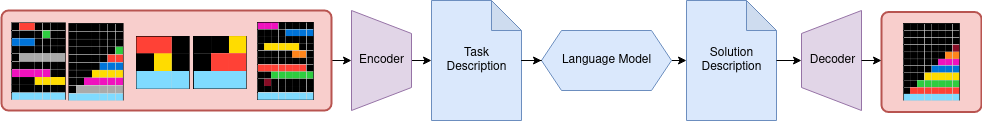}
        \end{center}
        \caption{The original tasks are given as image pairs in the visual domain (red) and are then transformed to the text domain (blue) using the encoder. The textual form of the task description allows a language model to propose a solution which is decoded back to the visual domain.}
        \label{fig:pipeline}
    \end{figure*}

    We believe that another promising path to include such priors is through a learning-based approach centered on natural language. 
    Natural language itself has evolved to incorporate human priors as an efficient way to describe our world.
    Furthermore, natural language has been shown to allow for better generalization across different tasks \cite{narasimhan2018grounding}.
    Recently, the emergence of Large Language Models (LLMs) has demonstrated success in understanding and producing natural language.
    Because these models have seen incredibly large amounts of human-generated text during their unsupervised pre-training, they have implicitly incorporated human priors.
    
    We motivate to leverage these insights and capabilities to automate the usage of language as a tool to create the foundation for a learning-centric approach to tackle ARC.
    We propose a general framework by first transforming an ARC task into a textual description of the entire task.
    A generic language model can then be used in a black-box fashion to output a description of the missing solution.
    Finally, the predicted solution can be translated back to the visual domain.
    An illustration of the entire framework is shown in Figure \ref{fig:pipeline}.
    Furthermore, we implement a first pipeline based on this framework to demonstrate the viability of this approach using a variety of available language models. In particular, through our learning-centric approach, we are able to solve tasks which could not be solved by any of the existing state-of-the-art DSL solutions.

\section{Related Work}

    The Abstraction and Reasoning Corpus, proposed by Chollet~\cite{chollet12019}, consists of 400 training tasks, 400 validation tasks and 200 tasks which are withheld for private evaluation.
    Each of the 800 publicly available tasks consists of two to six image pairs.
    Given these few example demonstrations, the goal is to come up with the missing solution of a specific test input image.
    Initially, the challenge was hosted as a Kaggle competition~\cite{ARCKaggle} and has recently seen a revival by Lab42~\cite{ARCLab42}.
    Many of the top performing solutions rely on an ensemble of heuristic program searches using hand-crafted domain specific languages \cite{IceCuber, Kaggle2, Kaggle3} and achieve up to 30 percent accuracy on the private evaluation set.
    However, there is still a large gap to match human performance, which is estimated to be at least around 80 percent~\cite{HumanARC}.
    Besides these approaches, there exist other proposals which incorporate some sort of learning based techniques such as neurosymbolic approaches~\cite{alfordARCDreamCoder}, graph representations~\cite{ViRel} or object-centric approaches~\cite{ARCGraph, assouel2022objectcentric}.
    In addition, recent efforts have started to investigate how natural language can help solve ARC tasks~\cite{LARC}. 
    However, they provide a human-generated text guidance on how to solve a task, whereas in this work we focus on describing and solving the complete task in an automated way in the language domain.
    \begin{table}[h]
            \centering
            \renewcommand{\arraystretch}{1.2}
            \begin{tabular}{|c|c|c|c|}
                \hline
                 \textbf{Model} & \textbf{Size} & \textbf{Layers} & \textbf{Hidden Dim.}  \\
                 \hline
                 bloom-560m~\cite{scao2022bloom} & $560$M & 24 & 1024\\
                 bloom-1b1~\cite{scao2022bloom}  & $1.1$B & 24 & 1536\\
                 bloom-1b7~\cite{scao2022bloom} & $1.7$B & 24 & 2048\\
                 bloom-3b~\cite{scao2022bloom} & $3$B & 30 & 2560\\
                 bloom-7b1~\cite{scao2022bloom} & $7.1$B & 30 & 4096\\
                 GPT-3~\cite{gpt3} & $175$B & 96 & 12228\\
                 \hline
            \end{tabular}
            \caption{Summary of the language models used in the experiments. The size of each model is given as number of parameters. }
            \label{tab:modelspecs}
        \end{table}
            \begin{table*}[t]
            \small
            \renewcommand{\arraystretch}{1.2}
            \noindent\makebox[\textwidth]{%
            \begin{tabular}{|c|c|c|c|c|c|c|}
                \hline
                \textbf{Rectangle} & \textbf{Line} & \textbf{Square} & \textbf{Pixel} & \textbf{Cross} & \textbf{Diagonal Cross} & \textbf{Complex}\\
                \cline{0-6}
                    & \includegraphics[scale=0.075]{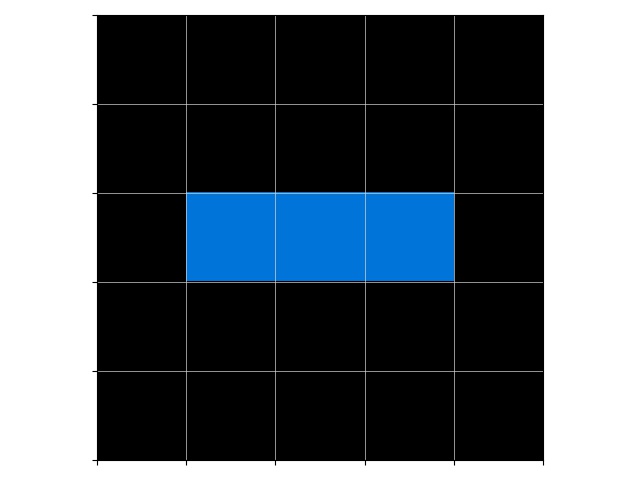}
                    & & & & & \rule{0pt}{40pt}\includegraphics[scale=0.075]{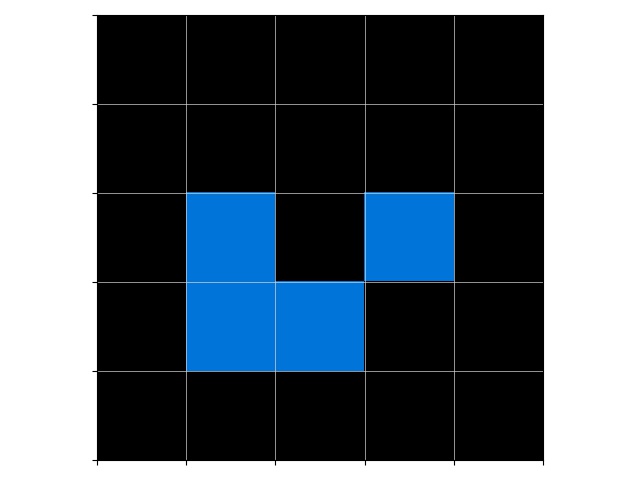}\\
                     \includegraphics[scale=0.075]{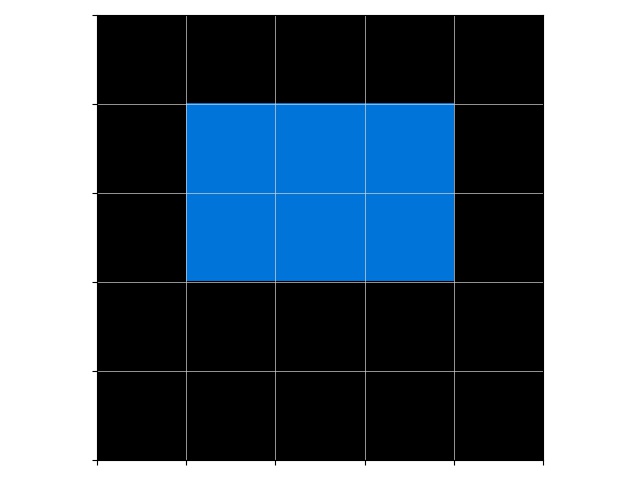} & 
                    \includegraphics[scale=0.075]{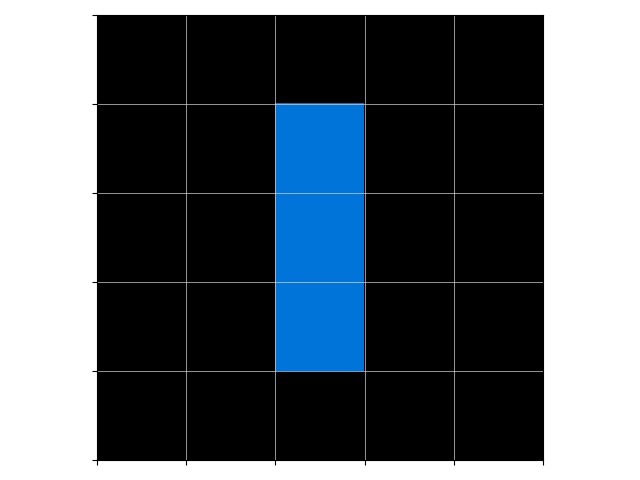} &
                    \includegraphics[scale=0.075]{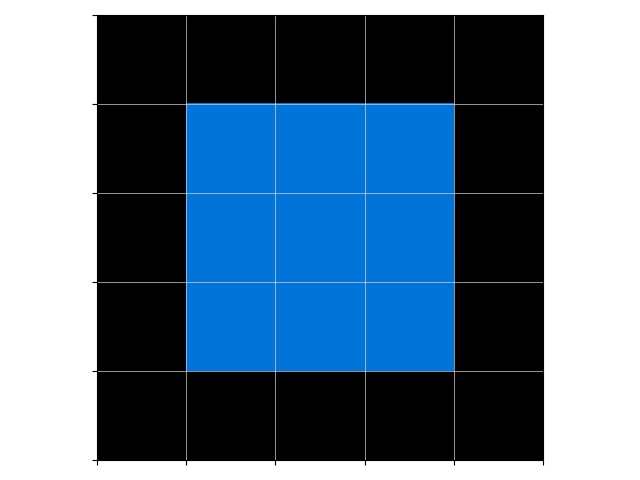} &
                    \includegraphics[scale=0.075]{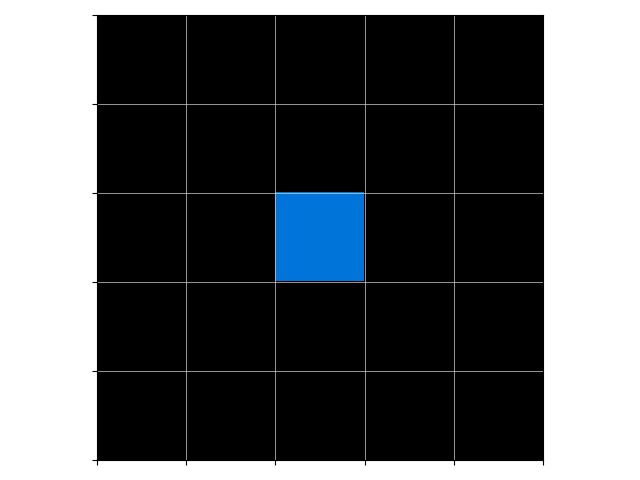} &
                    \includegraphics[scale=0.075]{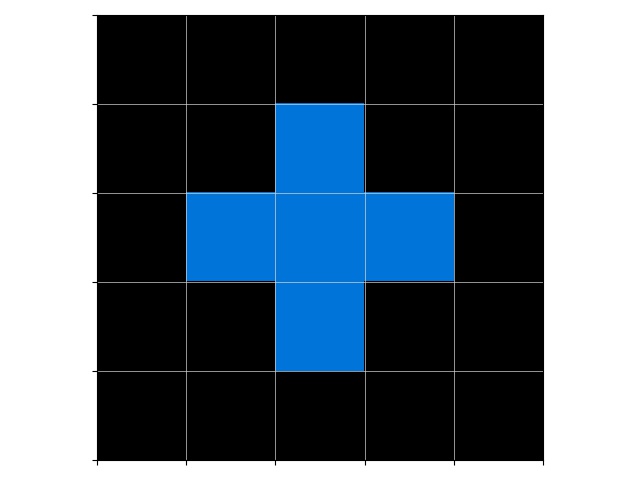} &
                    \includegraphics[scale=0.075]{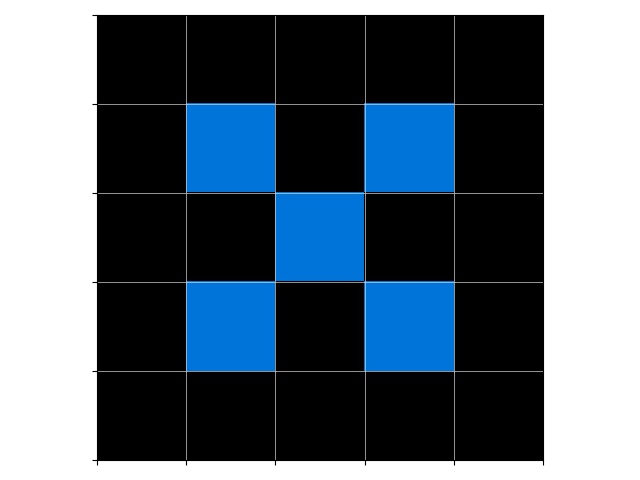} &
                    \includegraphics[scale=0.075]{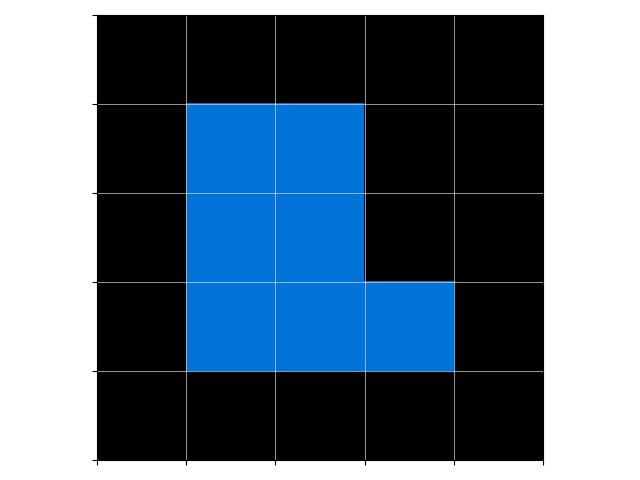} \\
                \hline
            \end{tabular}}
            \caption{Different types of objects that the encoder recognizes with their corresponding monochromatic blue image. Complex objects could capture any objects that do not fit any of the other categories, here two examples are shown. Complex objects are assigned individual IDs which are used consistently across textual descriptions of different images within a task.}
            \label{table:objetcs}
        \end{table*}

\section{Approach}
    We propose a learning-centric approach for the ARC challenge, which relies on a pipeline that can utilize unsupervised pre-trained models. The primary goal is to enable the learning of important priors, such as goal-directedness, numbers and counting, rather than requiring the model designer to encode them manually.
    Our prototype uses a modular design, decomposing the language and vision components. Currently, the vision module utilizes a heuristic approach, while the language module relies on pre-trained language models. We leave the development of a learning-based vision module to future work and focus on this specific setting to demonstrate the potential of tackling ARC using this pipeline to combine vision and natural language.
    
    \begin{figure}
                \texttt{\small \noindent \textbf{Input 1:} 10x10 grid, black background. Objects: continuous random object, of shape 'B' with symmetric shape along both axis with upper left corner in position (2,1), of size 3x4, monochromatic of color cyan. continuous random object, of shape 'B' with symmetric shape along both axis with upper left corner in position (7,5), of size 3x4, monochromatic of color cyan. continuous random object, of shape 'A' with symmetric shape along both axis with upper left corner in position (1,7), of size 3x3, monochromatic of color cyan.}
                \small \newline
                \texttt{\small \noindent \textbf{Output 1:} 10x10 grid, black background. Objects: continuous random object, of shape 'B' with symmetric shape along both axis with upper left corner in position (2,1), of size 3x4, monochromatic of color blue. continuous random object, of shape 'B' with symmetric shape along both axis with upper left corner in position (7,5), of size 3x4, monochromatic of color blue. continuous random object, of shape 'A' with symmetric shape along both axis with upper left corner in position (1,7), of size 3x3, monochromatic of color red.}
                \caption{Textual description for the first input output pair of the task shown in Figure \ref{fig:arc_task}. The complete task description includes all pairs in combination with the input for the test image. The missing output has to be completed by the language model.}
                \label{fig:text_descr}
            \end{figure}
    
    \textbf{Vision Module}
        The vision module in our approach uses a fixed heuristic based on human visual priors, including objectness, basic geometry, and symmetry. The module consists of an encoder and decoder, extracting different objects present in the task images and describing them in natural language. The inverse rules of encoding are applied to decode a description.
        First, the encoder checks if the full input image appears as a subset of the output and vice versa, taking into account different rotations, reflections and scales. In this case, the whole image is regarded as a single complex object.
        Otherwise, it predicts the background color using a convolutional neural network (CNN) and retrieves all objects present in the images (see Table \ref{table:objetcs}), by detecting monochromatic or multichromatic contiguous sets of pixels different from the background, independent of scale.
        To derive the textual description, the encoder first describes the size of the grid and the background color.
        It then groups all objects by similar shape and color patterns, and orders them by size to generate the description. 
        The encoder and decoder could be replaced with learned modules, for example a pre-trained image captioner.
    
    \textbf{Language Module}
        The language module receives a textual description for each task consisting of all input output pairs as well as the input test image. The description of the missing output image has to be predicted by the language model. 
        We can use any pre-trained generative language model in a black-box fashion as part of our pipeline.
        We summarize all evaluated language models in Table \ref{tab:modelspecs}. Throughout our evaluation, none of the used models were fine-tuned and have not seen ARC descriptions. Note, that even though multiple input output image descriptions are provided, the language model has no access to entire previous task descriptions which makes this a zero-shot setting.
        We include minimal pre- and post-processing steps of the textual descriptions such as trimming trailing text and skipping lenghty descriptions.
        To reduce variance resulting from the stochastic text generation process of the language models, we allowed 10 guesses per model and count a task as solved if at least 1 of the guesses is correct.

    \begin{figure*}[t]
        \begin{subfigure}[b]{0.5\textwidth}
             \centering       
             \includegraphics[width=\linewidth]{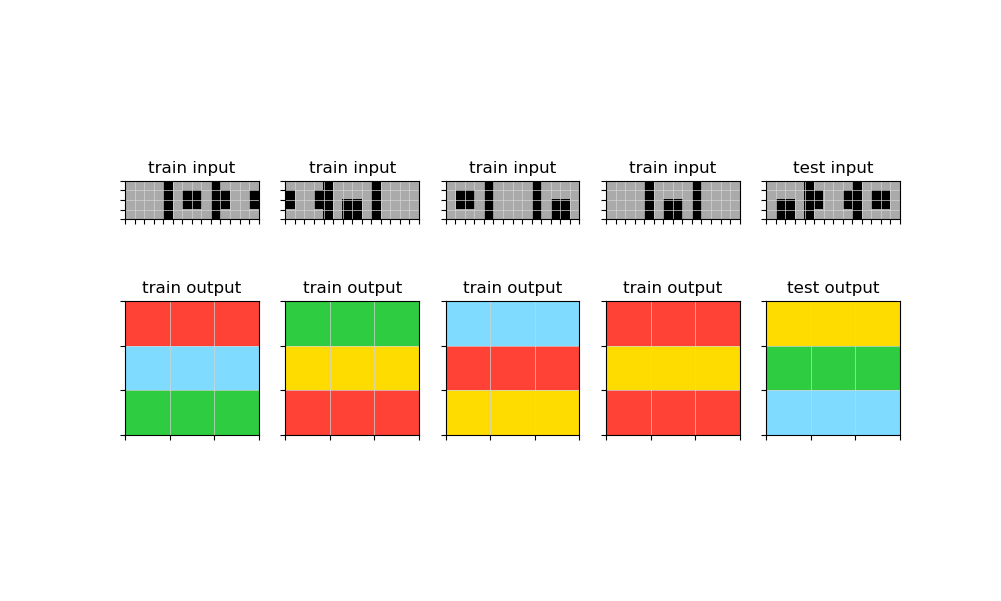}
             \caption{995c5fa3.json}
         \end{subfigure}
         \begin{subfigure}[b]{0.5\textwidth}
             \centering
             \includegraphics[width=\linewidth]{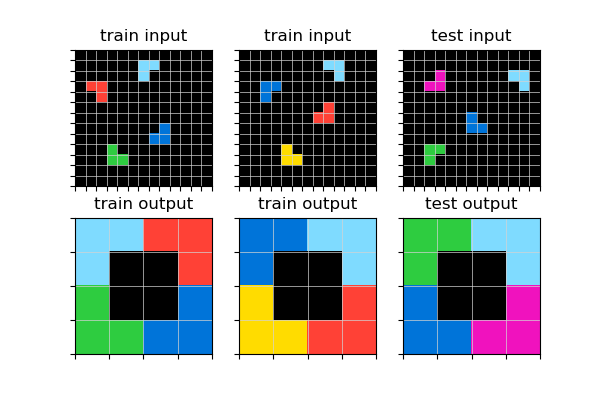}
             \caption{a61ba2ce.json}
         \end{subfigure}
         \newline
         \begin{subfigure}[b]{0.5\textwidth}
             \centering
             \includegraphics[width=\linewidth]{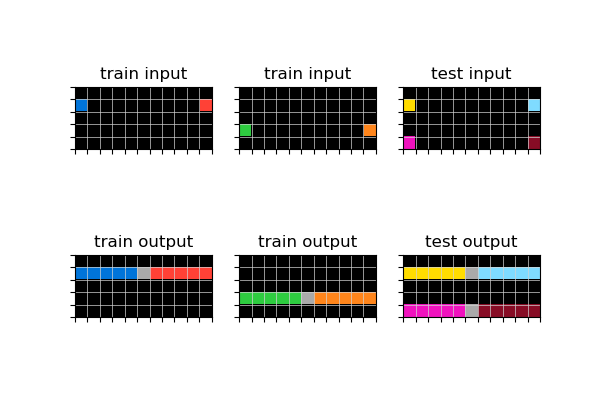}
             \caption{29c11459.json}
         \end{subfigure}
         \begin{subfigure}[b]{0.5\textwidth}
             \centering
             \includegraphics[width=\linewidth]{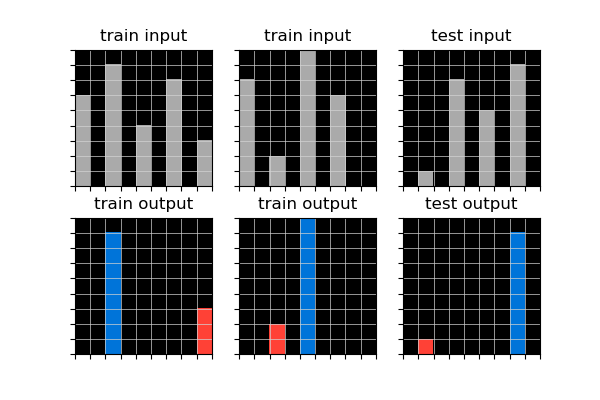}
             \caption{a61f2674.json}
         \end{subfigure}
        \caption{A collection of ARC tasks that were solved using our approach but not by the top performing solutions\cite{IceCuber, Kaggle2, Kaggle3}}
        \label{fig:betterthan}
    \end{figure*}
    
\section{Results}
    To empirically evaluate our framework and implemented pipeline, we carried out experiments using various language models of different sizes of the open-source Bloom architecture \cite{scao2022bloom} as well as GPT-3 \cite{gpt3}, namely \texttt{text-davinci-003}.
                
\begin{figure}[H]
            \begin{center}
            \small
            \includegraphics[width=8.5cm, height=5.5cm]{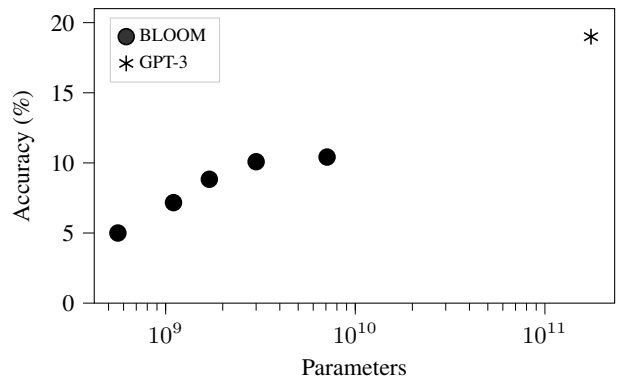}
            \vspace{-10pt}
            \end{center}
            \caption{Accuracy of our approach using language models of different sizes on the ARC training set. The accuracy follows a log-improvement as the number of parameters increases.}
            \label{fig:results}
        \end{figure}

    Note that for all our experiments, we generate the solutions in a zero-shot setting and did not fine-tune any of the language models. 
    We evaluate our pipeline on the ARC training set and report the accuracy in Figure \ref{fig:results}. 
    Without ever encountering an ARC task during training, even the smallest language models can solve a non-trivial amount of tasks. However, most notably, the accuracy  increases roughly linear with the logarithm of the number of parameters of the LLM. 

    Therefore, using our proposed framework it is possible to solve entire ARC tasks in an automated fashion that leverages pre-trained learned models.
    Moreover, larger models seem to perform better which hints at a potential increase in their capabilities. 
    However, another key aspect of this approach compared to hand-crafted program searches is that the priors can be implicitly learned and are not fixed by the designer.
    This has the advantage that we could solve novel tasks which cannot be captured by existing approaches, as they do not match their specified priors exactly.
    We compare the set of successfully solved tasks of our approach with the top performing solutions of the original Kaggle competition \cite{IceCuber, Kaggle2, Kaggle3}.
    We find that using our approach, we were able to solve tasks which could not be solved by any of the mentioned top performing solutions.
    A selection of such newly solved examples is shown in Figure \ref{fig:betterthan}.

\section{Conclusion}
    Current learning based approaches to solve ARC fail to be competitive, one reason being the limited number of tasks. We motivate to focus on natural language as a source of important human aligned priors to alleviate this lack of data. 
    We present a general framework to combine visual abstraction and leverage language as a reasoning tool in order to tackle ARC.
    Furthermore, we implement a pipeline to solve tasks in an automatic end-to-end fashion based on large language models as zero-shot reasoners.
    The obtained results are a promising demonstration of the framework and indicate the viability of our approach.
    In particular, we were able to solve new tasks compared to state-of-the-art DSL approaches.
    Moreover, we highlight that there exist many possibilities for extensions such as a learned encoding or decoding step, fine-tuned LLMs or even end-to-end training of the entire pipeline.




{\small
\bibliographystyle{ieee_fullname}
\bibliography{egbib}
}

\end{document}